\newif\ifarxiv
\newif\ifral
\newif\ifralfinal

\ralfalse \ralfinalfalse \arxivtrue %

\ifarxiv
\documentclass[letterpaper, 10 pt, conference]{ieeeconf}  %
\fi
\ifral
\documentclass[letterpaper, 10 pt, conference]{ieeeconf}  %
\fi
\ifralfinal
\documentclass[letterpaper, 10 pt, journal, twoside]{IEEEtran}
\fi

\IEEEoverridecommandlockouts                              %

\usepackage{booktabs}
\usepackage{graphics} %
\usepackage{epsfig} %
\usepackage{mathptmx} %
\usepackage{times} %
\usepackage{amsmath} %
\usepackage{amssymb}  %
\usepackage{arydshln} %
\usepackage{capt-of}
\usepackage{textcomp}  %
\usepackage[dvipsnames]{xcolor}
\usepackage{xcolor,colortbl}
\usepackage[accsupp]{axessibility}  %
\usepackage{cite}
\usepackage{multirow}
\usepackage{array}
\let\labelindent\relax
\usepackage[shortlabels]{enumitem} %
\makeatletter
\let\NAT@parse\undefined
\makeatother
\definecolor{citepurple}{rgb}{0.288,0.1196,0.7}
\usepackage[breaklinks,colorlinks,bookmarks,citecolor=citepurple]{hyperref}

\definecolor{Gray}{gray}{0.90}
\newcolumntype{g}{>{\columncolor{Gray}}c}
\definecolor{ffe1da}{RGB}{255,225,218}
\definecolor{F7E0D5}{RGB}{247,224,213}
\definecolor{darkF7E0D5}{RGB}{209,154,128}
\colorlet{Light}{White!0!F7E0D5}

\definecolor{OutdoorDark}{rgb}{0,.5,0}
\definecolor{IndoorDark}{rgb}{0,0.3,0.8}
\definecolor{SubTDark}{rgb}{0.5,.27,0.11}
\definecolor{AerialDark}{rgb}{.5,.0,.5}
\definecolor{UnderWaterDark}{rgb}{0.16, 0.46, 0.81}
\colorlet{Outdoor}{OutdoorDark!20!white}
\colorlet{Indoor}{IndoorDark!20!white}
\colorlet{SubT}{SubTDark!20!white}
\colorlet{Aerial}{AerialDark!20!white}
\colorlet{UnderWater}{UnderWaterDark!20!white}
\colorlet{OutdoorLight}{OutdoorDark!70!white}
\colorlet{IndoorLight}{IndoorDark!70!white}
\colorlet{SubTLight}{SubTDark!70!white}
\colorlet{AerialLight}{AerialDark!70!white}
\colorlet{UnderWaterLight}{UnderWaterDark!70!white}
\newcommand{\symbolHt}{1.5em}
\newcommand{\outdoorChar}{%
  \begingroup\normalfont
  \includegraphics[height=\symbolHt]{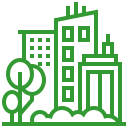}%
  \endgroup
}
\newcommand{\indoorChar}{%
  \begingroup\normalfont
  \includegraphics[height=\symbolHt]{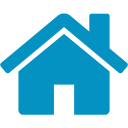}%
  \endgroup
}
\newcommand{\subtChar}{%
  \begingroup\normalfont
  \includegraphics[height=\symbolHt]{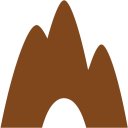}%
  \endgroup
}
\newcommand{\hawkinsChar}{%
  \begingroup\normalfont
  \includegraphics[height=\symbolHt]{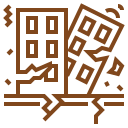}%
  \endgroup
}
\newcommand{\aerialChar}{%
  \begingroup\normalfont
  \includegraphics[height=\symbolHt]{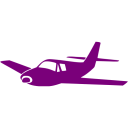}%
  \endgroup
}
\newcommand{\underwaterChar}{%
  \begingroup\normalfont
  \includegraphics[height=\symbolHt]{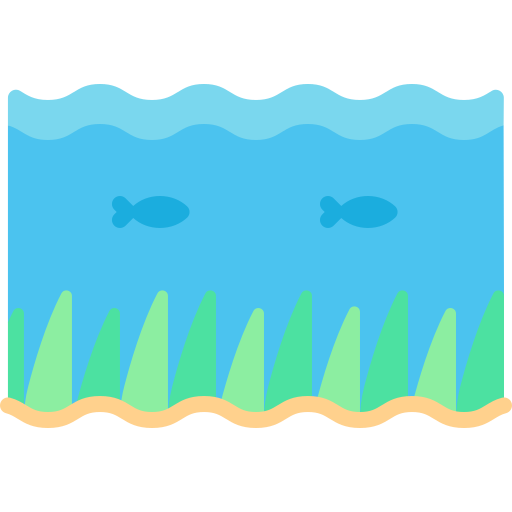}%
  \endgroup
}
\newcommand{\viewpointChar}{%
  \begingroup\normalfont
  \includegraphics[height=\symbolHt]{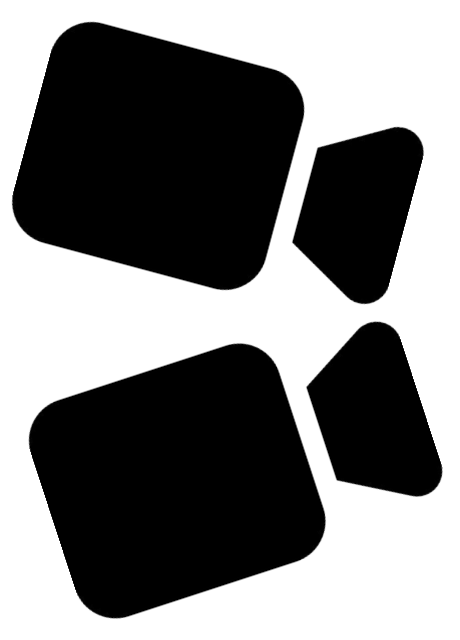}%
  \endgroup
}
\newcommand{\lightingChar}{%
  \begingroup\normalfont
  \includegraphics[height=\symbolHt]{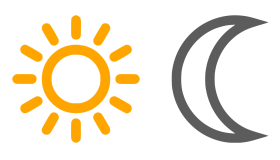}%
  \endgroup
}
\newcommand{\oppsymbolHt}{1em}
\newcommand{\oppositeChar}{%
  \begingroup\normalfont
  \includegraphics[height=\oppsymbolHt]{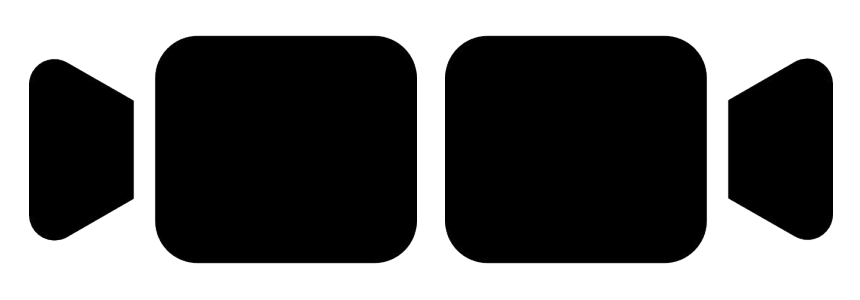}%
  \endgroup
}

\newcommand{\urban}[1]{\textbf{\textcolor{OutdoorDark}{Urban}}}
\newcommand{\indoor}[1]{\textbf{\textcolor{IndoorDark}{Indoor}}}
\newcommand{\aerial}[1]{\textbf{\textcolor{AerialDark}{Aerial}}}
\newcommand{\subt}[1]{\textbf{\textcolor{SubTDark}{SubT}}}
\newcommand{\degraded}[1]{\textbf{\textcolor{SubTDark}{Degraded}}}
\newcommand{\underwater}[1]{\textbf{\textcolor{UnderWaterDark}{Underwater}}}

\definecolor{darkpurple}{rgb}{0.288,0.1196,0.7}

\definecolor{amber}{rgb}{1.0, 0.75, 0.0}

\newcommand{\Rom}[1]{\uppercase\expandafter{\romannumeral #1\relax}}

\definecolor{darkgray}{rgb}{0.2, 0.2, 0.2}
\newcommand{\highlight}[1]{\textcolor{darkgray}{\textbf{#1}}}

\usepackage[capitalize]{cleveref}
\crefname{section}{Section}{Sections}
\crefname{table}{Table}{Tables}

\makeatletter
\newcommand{\cdashmidrule}[1]{%
  \noalign{\vskip\aboverulesep}
  \cdashline{#1}
  \noalign{\vskip\belowrulesep}}
\makeatother

\newcommand*{\subfigref}[2][]{%
  Fig. \hyperref[{fig:#2}]{%
    \ref*{fig:#2}%
    \ifx\\#1\\%
    \else
      \,#1%
    \fi
  }%
}

\newcommand{\coolname}{\textit{AnyLoc}}
\newcommand{\coolagg}[2]{\textit{AnyLoc-{#1}-{#2}}}
\newcommand{\coolaggshort}[1]{\textit{AnyLoc-{#1}}}
\newcommand{\dino}{\mbox{DINO}}
\newcommand{\dinovtwo}{\mbox{DINOv2}}
\definecolor{darkorange}{rgb}{1.0, 0.54, 0}
\newcommand{\webpage}{https://anyloc.github.io/}

\newcommand{\authorhref}[3][citepurple]{\href{#2}{\color{#1}{#3}}}

\title{\LARGE \bf
\coolname: Towards Universal Visual Place Recognition \\
\Large{\href{\webpage}{\color{orange}{\webpage}}}
}

\author{
\authorhref{https://nik-v9.github.io/}{Nikhil Keetha}$^{*1}$, 
\authorhref{https://theprojectsguy.github.io/}{Avneesh Mishra}$^{*2}$, 
\authorhref{https://jaykarhade.github.io/}{Jay Karhade}$^{*1}$, 
\authorhref{https://krrish94.github.io/}{Krishna Murthy Jatavallabhula}$^{3}$,
\\
\authorhref{https://theairlab.org/team/sebastian/}{Sebastian Scherer}$^{1}$, 
\authorhref{https://robotics.iiit.ac.in/faculty_mkrishna/}{Madhava Krishna}$^{2}$, and 
\authorhref{https://researchers.adelaide.edu.au/profile/sourav.garg}{Sourav Garg}$^{4}$
\\[5 pt]
$^{1}$\href{https://www.ri.cmu.edu/}{CMU}, 
$^{2}$\href{https://robotics.iiit.ac.in//}{IIIT Hyderabad}, 
$^{3}$\href{https://www.csail.mit.edu/}{MIT}, 
$^{4}$\href{https://www.adelaide.edu.au/aiml/}{University of Adelaide}
\thanks{*Equal Contribution}%
}

\begin{document}

\ifral
\bstctlcite{IEEEexample:BSTcontrol}
\fi
\ifralfinal
\bstctlcite{IEEEexample:BSTcontrol}
\fi

\makeatletter
\let\@oldmaketitle\@maketitle
\renewcommand{\@maketitle}{\@oldmaketitle
\centering
\begin{tabular}{cccc}
\includegraphics[trim={0cm 9.7cm 0cm 0cm},clip,width=0.96\linewidth]{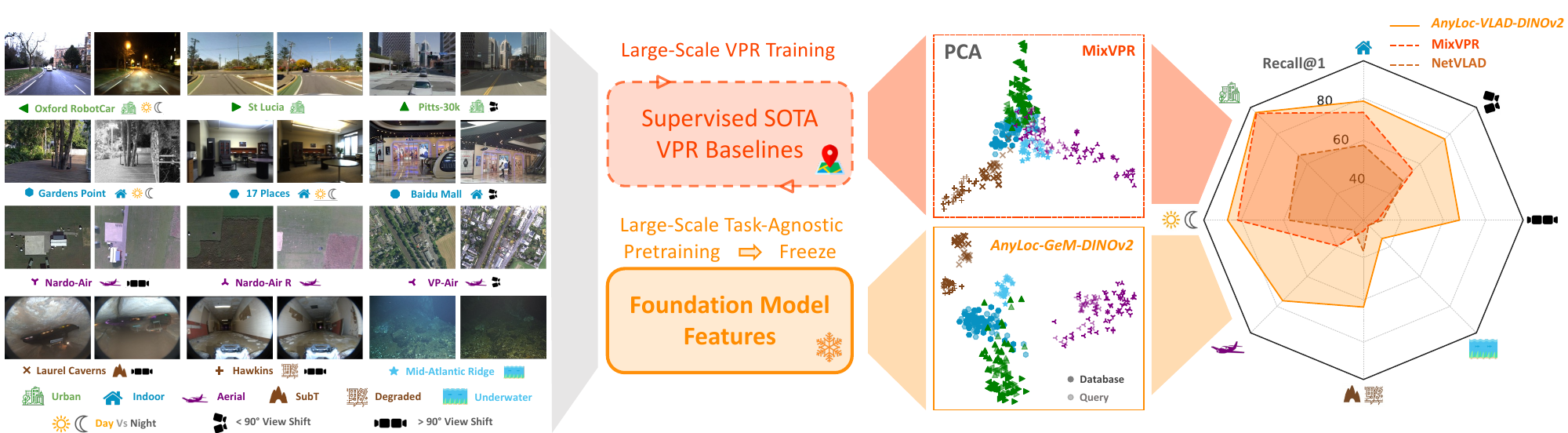}
\end{tabular}
\captionof{figure}{
\textbf{\coolname{}} enables \emph{universal} visual place recognition (VPR) across a massively diverse set of environments (\textit{anywhere}), temporal changes (\textit{anytime}), and a wide range of viewpoint variations (\textit{anyview}).
\coolname{} achieves this by aggregating per-pixel features extracted from large-scale pretrained models (\emph{foundation models}), \emph{without any training or finetuning}.
In the PCA panels (\emph{middle)}, notice how the features from MixVPR --- a state-of-the-art method trained specifically for VPR --- concentrate to a small region of the feature space, losing discriminative ability. On the other hand, \coolname{} uncovers 
distinct \textit{domains} encompassing datasets with similar properties, marked with the same color.
Using these \textit{domains} to construct vocabularies for unsupervised VLAD aggregation enables \coolname{} to achieve up to $4\times$ higher Recall@1, as seen in the polygonal areas in the radar chart (\textit{right}), across structured (urban outdoors, indoors) and unstructured (underwater, aerial, subterranean, visually degraded) environments.
}
\label{fig:splash}
}
\makeatother

\maketitle
\thispagestyle{empty}
\pagestyle{empty}

\begin{abstract}

Visual Place Recognition (VPR) is vital for robot localization.
To date, the most performant VPR approaches are \emph{environment- and task-specific}: while they exhibit strong performance in structured environments (predominantly urban driving), their performance degrades severely in unstructured environments, rendering most approaches brittle to robust real-world deployment.
In this work, we develop a \emph{universal} solution to VPR -- a technique that works across a broad range of structured and unstructured environments (urban, outdoors, indoors, aerial, underwater, and subterranean environments) without any re-training or finetuning.
We demonstrate that general-purpose feature representations derived from off-the-shelf self-supervised models \emph{with no VPR-specific training} are the right substrate upon which to build such a universal VPR solution.
Combining these derived features with \emph{unsupervised feature aggregation} enables our suite of methods, \coolname{}, to achieve up to $4\times$ significantly higher performance than existing approaches.
We further obtain a 6\% improvement in performance by characterizing the semantic properties of these features, uncovering unique \textit{domains} which encapsulate datasets from similar environments.
Our detailed experiments and analysis lay a foundation for building VPR solutions that may be deployed \emph{anywhere}, \emph{anytime}, and across \emph{anyview}.
We encourage the readers to explore our project page and interactive demos: \href{\webpage}{\color{orange}{\webpage}}.

\end{abstract}

\setcounter{figure}{1} %

\section{Introduction}
\label{sec:intro}

Visual Place Recognition (VPR) is a fundamental capability for robot state estimation and is widely applied in robotic systems such as autonomous cars, other uncrewed (aerial, terrestrial, and underwater) vehicles, and wearable devices. Despite significant advancements in VPR over the years, achieving out-of-the-box applicability across a diverse set of scenarios remains challenging; this is critical to bootstrap a mobile robot \textit{anywhere}, \textit{anytime}, and across \textit{anyview}.

State-of-the-art (SOTA) approaches are \emph{specifically trained} for VPR and exhibit strong performance on environments similar to those found in the training dataset (for instance, urban driving).
However, when the same methods are deployed in an environment where the extracted visual features differ substantially (such as underwater or aerial), their performance drops sharply~(\cref{fig:splash}).
In this context, we address the question, ``\textbf{How can one design a universal VPR solution?}"
This entails generating place representations from a \textit{general} model, which is pre-trained in an embodiment-, task- and environment-agnostic manner and can be readily adjusted to its \textit{specific} deployment environment.
Specifically, a \emph{universal} VPR solution must be applicable \emph{anywhere} (seamlessly operates across any environment, including aerial, subterranean, and underwater), \emph{anytime} (robust to temporal changes in the scene, such as day-night or seasonal variations, or to transient objects), and across \emph{anyview} (robust to perspective viewpoint variations, including diametrically opposite views).

We rethink the VPR problem from the lens of (visual) feature representations derived from large-scale pretrained models (coined \textit{foundation models}~\cite{fmodels}).
We show that, despite not being trained for VPR, these models encode rich visual features that serve as the right substrate upon which a \emph{universal} VPR solution may be built.
Our approach, termed \textbf{\coolname}, involves a careful selection of models and visual features with the \emph{right} invariance properties and blends them with prevailing local-aggregation approaches in the VPR literature~\cite{arandjelovic2016netvlad,berton2022deep,garrido2023on,shekhar2023objectives}, resulting in all of the aforementioned desirable characteristics of a \emph{universal} VPR solution.

\vspace{1em}

Our key takeaways are as follows:
\begin{itemize}
    \item \textit{AnyLoc} emerges as a new baseline VPR method that works universally across 12 datasets exhibiting massive diversity along the axes of \textit{place}, \textit{time}, and \textit{perspective};
    \item Self-supervised features (such as DINOv2~\cite{oquab2023dinov2}) and unsupervised aggregation methods (like VLAD~\cite{jegou2010aggregating} \& GeM~\cite{radenovic2018fine}) are \emph{both} crucial for strong VPR performance. Applying these aggregation techniques on per-pixel features offers substantial performance gains over the direct use of per-image features from off-the-shelf models.
    \item Characterizing the semantic properties of the aggregated local features uncovers distinct \emph{domains} in the latent space, which can further be used to enhance VLAD vocabulary construction; in turn boosting performance.
\end{itemize}

We evaluate \coolname{} on an extensive and diverse range of datasets (urban, indoors, aerial, underwater, subterranean) across challenging VPR conditions (day-night and seasonal variations, opposing viewpoints), establishing a strong baseline for future research towards universal VPR solutions.

\section{VPR: Overview, Trends \& Limitations}
\label{sec:state_of_vpr}

\textbf{VPR -- Problem definition}: VPR is often cast as an image retrieval problem~\cite{garg2021your} that comprises two phases. In the \emph{indexing} phase, a \highlight{reference map (image database)} is gathered from a robot's onboard camera when traversing through an environment. In the \emph{retrieval} phase, given a \highlight{query image}---captured during a future traverse---VPR entails retrieving the closest match to this query image in the reference map. There exists a variety of VPR methods and alternative problem formulations~\cite{lowry2015visual,berton2022deep,pion2020benchmarking,zaffar2021vpr,schubert2023visual}; in this work,
we focus on \highlight{global descriptors}, which offer the best tradeoff between accurate matching and search efficiency~\cite{jegou2010aggregating,sattler2018benchmarking,garg2021your}. This is in contrast to local descriptor methods, which are computationally intensive to match, particularly over larger databases.

\ifarxiv
Researchers have explored various training objectives~\cite{ge2020self,xiao2023visual,leyva2023data,berton2022rethinking}, aggregation techniques~\cite{arandjelovic2016netvlad,radenovic2018fine,chen2021learning}, and transfer learning~\cite{haas2023learning,berton2021adaptive,latif2018addressing} to improve global descriptor-based VPR. 
\fi
\ifral
Researchers have explored various training objectives~\cite{ge2020self,xiao2023visual,leyva2023data,berton2022rethinking}, aggregation techniques~\cite{arandjelovic2016netvlad,radenovic2018fine}, and transfer learning~\cite{haas2023learning,berton2021adaptive,latif2018addressing} to improve global descriptor-based VPR.
\fi
\ifralfinal
Researchers have explored various training objectives~\cite{ge2020self,xiao2023visual,leyva2023data,berton2022rethinking}, aggregation techniques~\cite{arandjelovic2016netvlad,radenovic2018fine}, and transfer learning~\cite{haas2023learning,berton2021adaptive,latif2018addressing} to improve global descriptor-based VPR.
\fi
High performance of most of these modern approaches can be attributed to \highlight{large-scale training on VPR-specific data}.
Powered by deep learning and the Pitts-250k dataset~\cite{torii2013visual}, weakly-supervised contrastive learning in NetVLAD~\cite{arandjelovic2016netvlad} led to substantial improvements over classical hand-crafted features. 
Following suit, the Google-Landmark V1 ($1$ million images) and V2 datasets~\cite{weyand2020google} ($5$ million images) enabled training DeLF~\cite{noh2017large} and DeLG~\cite{cao2020unifying} for large-scale image retrieval.
Likewise, the Mapillary Street-Level Sequences (MSLS) dataset, containing $1.6$ million \textit{street} images, substantially boosted VPR performance by tapping orders of magnitude larger data from urban and suburban settings~\cite{warburg2020mapillary, wang2022transvpr, zhu2023r2former}.
More recently, CosPlace~\cite{berton2022rethinking} 
coupled classification-based learning with the San Francisco XL dataset comprising $40$ million images having GPS \& heading.
The current SOTA, MixVPR~\cite{ali2023mixvpr}, proposed an MLP-based feature mixer, trained
on the GSV-Cities dataset~\cite{ali2022gsv} -- a curated large-scale dataset with 530,000 images spanning 62,000 places worldwide.

This trend of scaling up VPR training is mostly driven by easily-available positioning data for outdoor environments, which leads to SOTA performance in urban settings, but \highlight{does not
generalize to indoor and unstructured environments.}
As shown in \cref{fig:splash}, the PCA projections of descriptors extracted by
SOTA methods concentrate to a narrow region in the feature space, diminishing their discriminative abilities in environments outside the training distribution.
Apart from environment-specificity, prior methods have tackled \textit{specific} challenges in isolation, such as extreme temporal variations in scene appearance~\cite{latif2018addressing,tang2020adversarial} and camera viewpoint~\cite{garg2018lost,gawel2018x}.
This data- and task-specificity of current VPR approaches limits their out-of-the-box applicability, which
may be mitigated by task-agnostic learning.
Hence, in this work, we analyze the design space of VPR using web-scale self-supervised visual representations and \highlight{develop a \emph{universal} solution that does not assume any VPR-specific training}.

\section{AnyLoc: Towards Universal VPR}
\label{sec:foundloc}

To the best of our knowledge, our approach, \coolname{}, is the first VPR solution that exhibits \textit{anywhere}, \textit{anytime}, and \textit{anyview} capabilities (see~\cref{fig:splash}).
\coolname{} is guided by two crucial insights (see~\cref{sec:results} for details) that emerged when exploring the design space of VPR solutions through the lens of foundation model features: 
(a) \textit{existing VPR solutions are task-specific and perform poorly when evaluated in environments outside the training distribution}; 
and 
(b) \textit{while per-pixel features from off-the-shelf foundation models~\cite{caron2021emerging,oquab2023dinov2,radford2021learning} demonstrate remarkable visual and semantic consistency~\cite{park2023self,shekhar2023objectives,jatavallabhula2023conceptfusion,amir2021deep},
the per-image features are suboptimal when used as-is for VPR}.
Therefore, a careful investigation is needed to transfer these per-pixel invariances to the image level for recognizing \textit{places}, where recent approaches in this direction are only limited to small-scale indoor settings or vision-language use-cases~\cite{mirjalili2023fm, kassab2023clip}.
In this context, for designing \coolname{}, we investigate the following questions:
\begin{enumerate}[A.]
\item What foundation models are best suited to VPR?
\item How do we extract VPR-suited local features from these general-purpose models?
\item How do we aggregate \emph{local} features to describe \emph{places}?
\item How to construct vocabularies that generalize across datasets?
\end{enumerate}

\subsection{Choice of Foundation Model}
\label{sec:fmodels}

There exist three broad classes of \highlight{self-supervised foundation models that extract task-agnostic visual features}: (a) joint embedding methods (DINO~\cite{caron2021emerging}, DINOv2~\cite{oquab2023dinov2}), (b) contrastive learning methods (CLIP~\cite{radford2021learning}), and (c) masked autoencoding approaches (MAE~\cite{he2022masked}).
Joint embedding methods need a stable training recipe; \dino{} is trained on ImageNet~\cite{deng2009imagenet} through global image-level self-supervision, while \dinovtwo{} is trained on a much larger, carefully-curated dataset with joint image-/token-level losses.
These methods offer the highest level of performance; followed by contrastive learned approaches like CLIP~\cite{radford2021learning}, which is trained on millions of \textit{aligned} image-text pairs.
In our initial experiments, we found all these models to perform better than MAE~\cite{he2022masked}, which only has token-level self-supervision. These findings are corroborated in~\cite{park2023self,shekhar2023objectives,oquab2023dinov2}, highlighting the benefits of learning long-range global patterns captured by joint embedding methods.
Therefore, \coolname{} employs \dino{} \& \dinovtwo{} vision transformers for extracting features.

\subsection{Choice of Features}
\label{sec:features}

\begin{figure}[!t]
\centering
\begin{tabular}{cccc}
\includegraphics[trim={0cm 3.4cm 0cm 2.7cm},clip,width=0.95\linewidth]{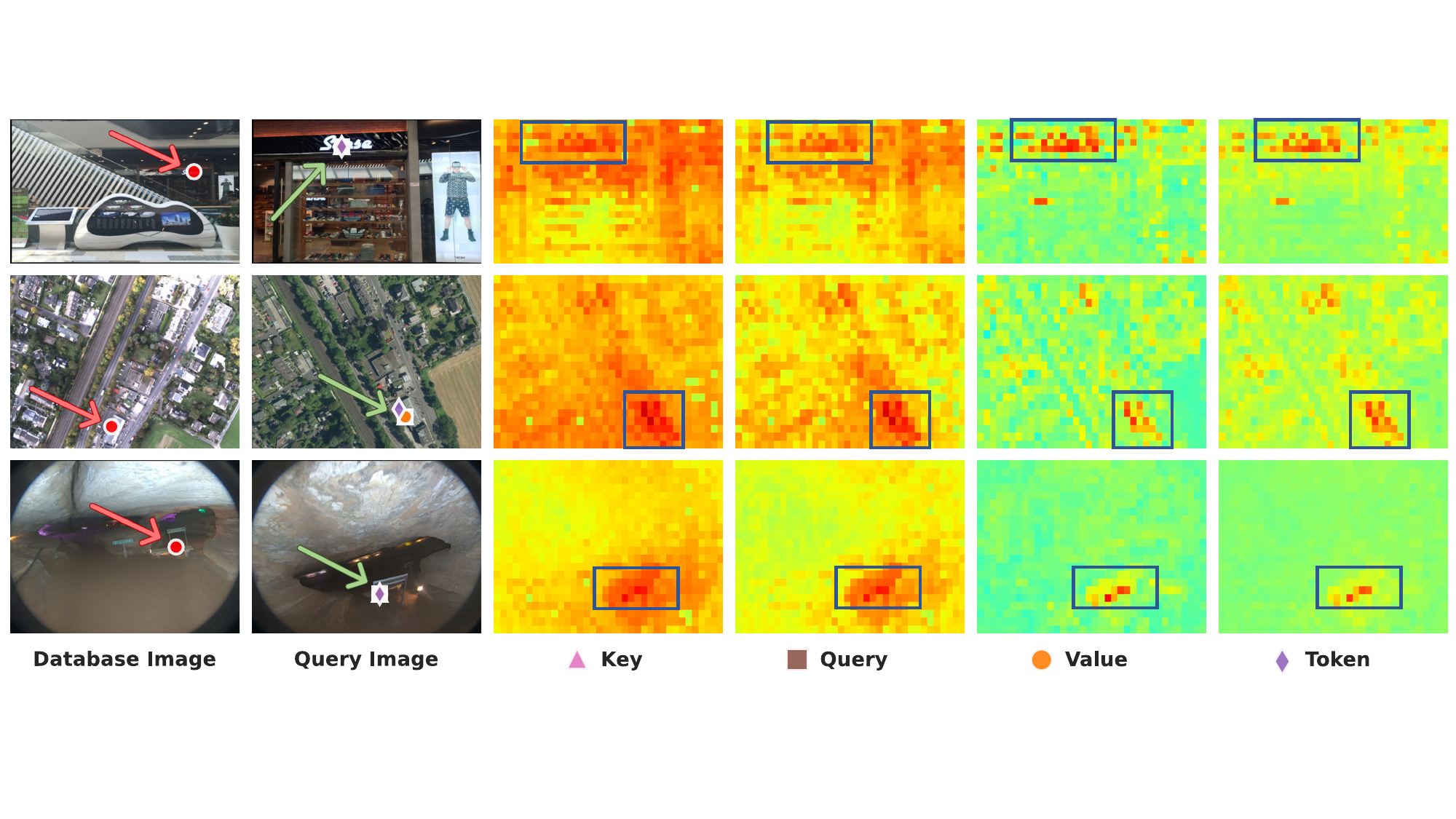} 
\end{tabular}
\caption{Point correspondences (as markers) \& similarity maps show the \textbf{robustness of foundation model features to various VPR challenges}: (\textit{top}) text and scale change, (\textit{middle}) perceptually aliased features and viewpoint shift, and (\textit{bottom}) low illumination combined with opposing viewpoint. The \texttt{value} facet has the highest contrast between the background and the matched points, which is vital for discarding distractors within an image.}
\label{fig:facet_sim}
\end{figure}

Another important design choice is how we extract visual features from these pretrained vision transformers (ViT)~\cite{dosovitskiy2020vit}.
Rather than extract per-image features\footnote{In a ViT, per-image features are encoded in a special token, \texttt{CLS}, and interpreted as a summary of the image content.}
(i.e., one feature vector for the entire image), we observe that per-pixel features enable fine-grained matching and result in superior performance.
Each layer in the ViT has multiple \emph{facets} (\texttt{query}, \texttt{key}, \texttt{value}, and \texttt{token}) from which features may be extracted.
Following~\cite{amir2021deep}, we extract features from intermediate layers across the ViT and discard the \texttt{CLS} token.
In~\cref{fig:facet_sim}, we illustrate this \highlight{applicability of the dense ViT features for VPR} by assessing the robustness of local feature correspondences.
We select a point on a database image, match it with all (per-pixel) features from the query image, and plot heatmaps indicating the likelihood these points correspond.
Notice how the correspondences are robust even in the presence of semantic text and scale change (first row), perceptual aliasing and viewpoint shift (second row), and low illumination combined with opposing viewpoint (third row).

\begin{figure}[!t]
\centering
\begin{tabular}{cccc}
\includegraphics[trim={0cm 2.05cm 0cm 1.9cm},clip,width=0.95\linewidth]{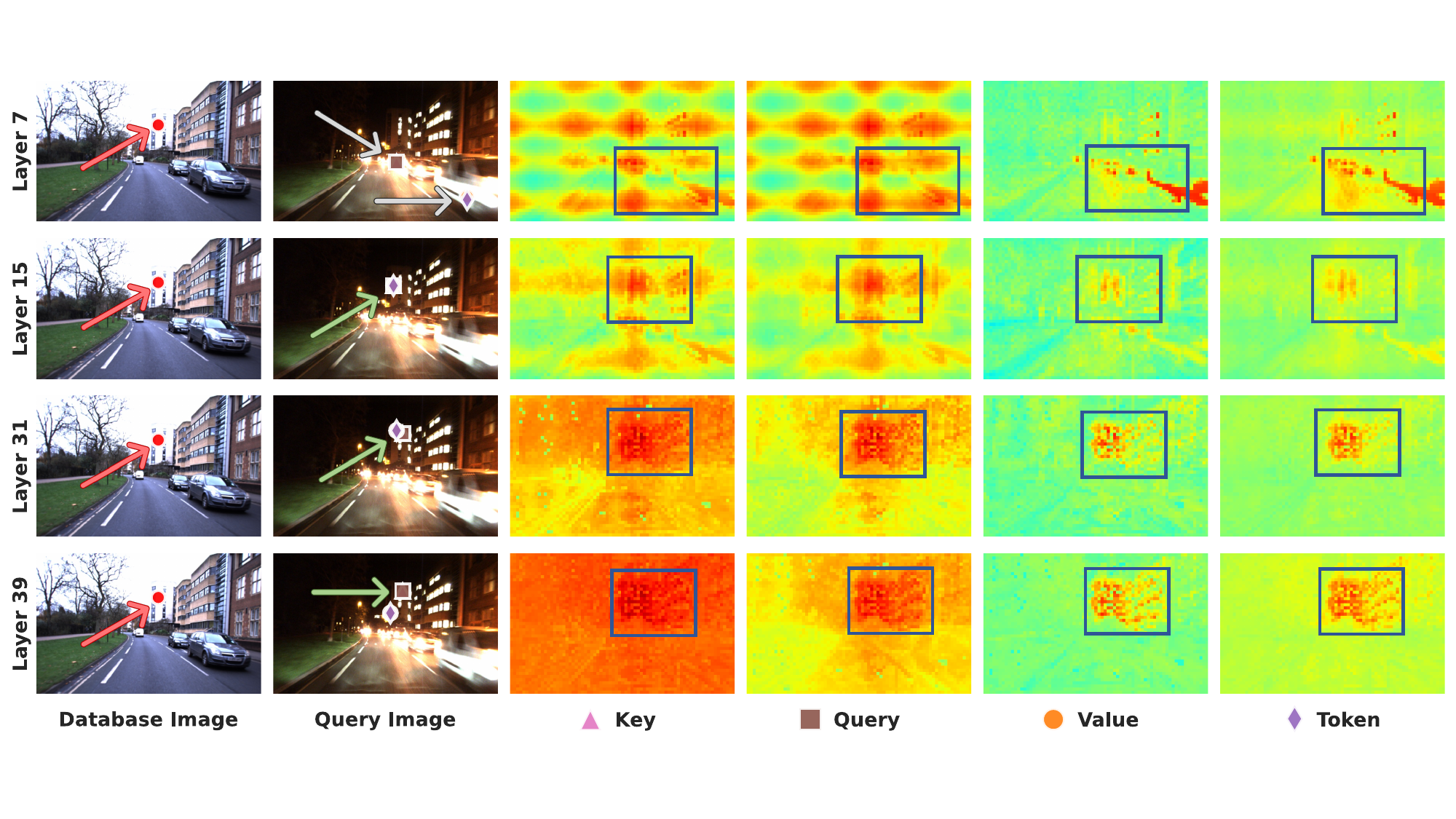} 
\end{tabular}
\caption{Qualitative ablation comparing the absolute-scale similarity maps of features from different \dinovtwo{} ViT-G \textit{layers} and \textit{facets}. \textbf{Layer 31 \textnormal{\texttt{value}} facet has the sharpest contrast} in the similarity map, which is \textbf{crucial for robustness against distractors within an image.}}
\label{fig:facet_sim_layer_ablation}
\end{figure}

Comparing the similarity maps in~\cref{fig:facet_sim}, notice how the \texttt{value} \highlight{facet exhibits the largest contrast}
between the matched points and the background, which is \highlight{crucial for robustness against distractors within an image}. 
Upon further analysis \textit{across layers} (\cref{fig:facet_sim_layer_ablation}), we observe an interesting trend. The earlier layers of the ViT (top rows), especially the \texttt{key} and \texttt{query} facets, exhibit a high positional encoding bias, while \highlight{the \textnormal{\texttt{value}} facet of deeper layers has the sharpest contrast} in the similarity map.
We further justify our selection of layer \& facet quantitatively in~\cref{sec:facets_and_layers}.

\subsection{Choice of Aggregation Technique}
\label{sec:aggregating_features}
The next design choice to make towards our VPR pipeline entails selecting an \emph{aggregation technique} that determines how local features are grouped together to describe sections of an image and, eventually, an environment.
While prior work has used the \texttt{CLS} token directly for image retrieval~\cite{el2021training, oquab2023dinov2, kassab2023clip},
we observed contradictory trends under a \emph{universal} retrieval setting (i.e., retraining or finetuning is prohibited).
We \highlight{comprehensively explore multiple unsupervised aggregation techniques}: Global Average Pooling (GAP)~\cite{babenko2015aggregating}, Global Max Pooling (GMP)~\cite{razavian2016visual}, Generalized Mean Pooling (GeM)~\cite{radenovic2018fine}, and the soft \& hard assignment variants of VLAD~\cite{jegou2010aggregating}.

For an input image of size $H \times W$, and a per-pixel feature $f_i \in \mathbb{R}^D$,
we define a global descriptor as: 
\begin{equation}
\centering
F_{G} = \left(\sum_{i = 1}^{H \times W} {f_i}^{p}\right)^{\frac{1}{p}}
\end{equation}
where $p = 1$, $p = 3$, and $p \to \infty$ represent GAP, GeM, and GMP respectively.

For VLAD variants, we cluster all the features from the database images to obtain $N$ cluster centers. This forms our \emph{vocabulary}.
The global VLAD descriptor is then calculated as the sum of residuals per cluster center $k$, as below:
\begin{equation}
\centering
F_{V_k} = \sum_{i=1}^{N \times H \times W} \alpha_{k}(f_{i})(f_{i} - c_{k})
\end{equation}
where $\alpha_{k}(x_i)$ is 1 if $f_i$ is assigned to cluster $k$ and 0 otherwise. 
In the soft-assignment variant of VLAD, $\alpha_{k}(f_i)$ indicates the assignment probability and lies between 0 and 1. Following~\cite{arandjelovic2013all}, we perform intra-normalization, concatenation, and inter-normalization to obtain the final VLAD descriptor $F_V$.

\subsection{Choice of Vocabulary}
\label{sec:vocab_approach}

For vocabulary-based aggregation techniques, we construct our vocabulary with the goal of \highlight{characterizing the distinct semantic properties of globally pooled local features across diverse environments}.
Prior work based on VLAD has either used a global vocabulary based on representative places \& features~\cite{jegou2010aggregating}, a reference map-specific one~\cite{arandjelovic2013all}, or a learnt~\cite{arandjelovic2016netvlad} vocabulary based on the training dataset. 
These approaches work well where domain- or map-specific data is abundant and task-specific training is feasible. 
However, a more scalable approach is to leverage the open-set semantic attributes encoded in the foundation model features to determine the appropriate domain and feature vocabulary. 
Hence, we use vocabulary-independent global descriptors (DINOv2-GeM) and their (unsupervised) PCA projection to define vocabularies for VLAD aggregation.

From~\cref{fig:splash}, we observe that \highlight{projecting the global descriptors using PCA uncovers distinct \emph{domains} in the latent space}, which characterizes datasets having similar properties, namely: \urban{}, \indoor{}, \aerial{}, \subt{}, \degraded{}, and \underwater{}.
Further demonstrating discriminative robustness, although the \subt{} and \degraded{} domains have similar imagery types, they are dispersed to distinct regions, whereas the visually degraded indoor domain is concentrated relatively close to the indoor collection.
Lastly, we can observe that the projected features for the query images are close to the projected features of their respective database images\footnote{The PCA transform is computed solely from the database images, and does not make use of the query images, for fair analysis.}.
Hence, using the PCA-based segregation, we construct the visual vocabularies for VLAD in a domain-specific manner (further justified in \cref{sec:vocab_ablations}).

\begin{table}[!t]
\centering
\caption{Unstructured Environments used in Evaluation}
\setlength\extrarowheight{-3pt}
\scalebox{0.85}{
\begin{tabular}{@{}lccccl@{}}
\toprule

\textbf{Dataset} & $\mathbf{N_{Db}}$ & $\mathbf{N_{Q}}$ & \textbf{Traj. Span} & \textbf{Loc. Radius} & \textbf{Type} \\

\cmidrule{1-1} \cmidrule(lr{0.75em}){2-2} \cmidrule(lr{0.75em}){3-3} \cmidrule(lr{0.75em}){4-4} \cmidrule(lr{0.75em}){5-5} \cmidrule{6-6}

\multirow{2}{*}{\color{SubTDark} \textbf{Hawkins}~\cite{zhao2023subtmrs}} & \multirow{2}{*}{65} & \multirow{2}{*}{101} & \multirow{2}{*}{282 m} & \multirow{2}{*}{8 m} & \multirow{2}{*}{\hawkinsChar} \\

&&&&& \\

\multirow{2}{*}{\color{SubTDark} \textbf{Laurel Caverns}~\cite{zhao2023subtmrs}} & \multirow{2}{*}{141} & \multirow{2}{*}{112} & \multirow{2}{*}{102 m} & \multirow{2}{*}{8 m} & \multirow{2}{*}{\subtChar} \\

&&&&& \\

\multirow{2}{*}{\color{AerialDark} \textbf{Nardo-Air}} & \multirow{2}{*}{102} & \multirow{2}{*}{71} & \multirow{2}{*}{700 m / 1 $km^2$} & \multirow{2}{*}{60 m} & \multirow{2}{*}{\aerialChar} \\

&&&&& \\

\multirow{2}{*}{\color{AerialDark} \textbf{VP-Air}~\cite{schleiss2022vpair}} & \multirow{2}{*}{12.7k} & \multirow{2}{*}{2.7k} & \multirow{2}{*}{100 km} & \multirow{2}{*}{3 frames} & \multirow{2}{*}{\aerialChar} \\

&&&&& \\

\multirow{2}{*}{\color{UnderWaterDark} \textbf{Mid-Atlantic Ridge}~\cite{boittiaux2022eiffel}} & \multirow{2}{*}{65} & \multirow{2}{*}{101} & \multirow{2}{*}{18 m} & \multirow{2}{*}{0.3 m} &  \multirow{2}{*}{\underwaterChar} \\

&&&&& \\

\bottomrule
\end{tabular}
}
\label{tab:datasets}
\end{table}

\section{Experimental Setup}
\label{sec:setup}

\subsection{Datasets}
\label{sec:datasets}

There exist several VPR datasets where the composition of benchmarks is influenced by either the end task, i.e., urban data for Geo-localization~\cite{berton2022deep} or the evaluation aspects of viewpoint variability~\cite{zaffar2021vpr}. 
We evaluate on datasets from both structured and unstructured environments, 
offering unprecedented diversity in terms of environments (\emph{anywhere}), coupled with a range of temporal (\emph{anytime}) and camera viewpoint\footnote{The viewpoint shifts range from $<90^\circ$ with minimal (Oxford, St Lucia) and moderate shifts (Pitts30K, Baidu) to $>90^\circ$ with extreme shifts (orthogonal in Nardo-Air and opposite in Hawkins, Laurel). $>90^\circ$ criterion for the opposite-viewpoint datasets refers to a $180^\circ$ orientation change in observing a place from a nearby but not the same 3D position~\cite{garg2018lost,garg2021your}.} (\emph{anyview}) variations.
We define structured environments as organized places composed of human-built structures that are commonplace in applications such as autonomous driving and indoor robotics. 
These represent the typical images collected and shared by humans on the web.
On the other hand, unstructured environments represent in-the-wild scenarios where the objects and types of images encountered are not commonly observed.

\subsubsection{Structured Environments}
\label{sec:structured_envs}

We used six \highlight{benchmark} indoor and outdoor datasets, exhibiting challenges like drastic viewpoint shifts, perceptual aliasing, and substantial visual appearance change. 
This includes Baidu Mall~\cite{sun2017baidu}, Gardens Point~\cite{glover2014gardens, sunderhauf2015performance}, 17 Places~\cite{sahdev2016indoor}, Pittsburgh-30k~\cite{torii2013visual}, St Lucia~\cite{warren2010unaided}, and Oxford RobotCar~\cite{maddern20171}, where the ground truth localization radius is $10$ meters, $2$ frames, $5$ frames, $25$ meters, $25$ meters, and $25$ meters, respectively. 
For Oxford RobotCar, we use a subsampled version of the Overcast Summer and Autumn Night traverses, following HEAPUtil~\cite{keetha2021hierarchical}.

\subsubsection{Unstructured Environments}
\label{sec:unstructured_envs}

While our structured environments enable us to benchmark with respect to existing VPR techniques, to \highlight{truly assess robustness and versatility}, we evaluate on a number of unstructured environments, including aerial, underwater, visually degraded, and subterranean environments\footnote{Models such as DINOv2 and CLIP are trained on web-scale datasets, and consequently, will likely have seen structured environments similar to those in Table~\ref{tab:mainResult}. Therefore, the true test for these models is their performance on unstructured environments, which are highly unlikely to have featured in any of the training subsets for these models.}.
\cref{tab:datasets} provides an overview of these unstructured datasets, 
which exhibit challenging distribution shifts, visually degraded long corridors, satellite \& aerial imagery covering various landscapes, low illumination, and seasonal variations.
Nardo-Air R aligns the orientation of drone imagery with the satellite map.

\subsection{Benchmarking \& Evaluation}
\label{sec:benchmarking}

We use Recall$@K$~\cite{zaffar2021vpr} as the evaluation metric (a higher recall score indicates superior performance). All experiments use the same random seed  (\texttt{42}) and GPU hardware (NVIDIA RTX $3090$) for consistency and reproducibility.

\subsubsection{State-of-the-art Baselines}
\label{sec:baselines}
We evaluate \coolname{} against a variety of VPR methods such that it encompasses variations in terms of VPR-specific training, global image representation, type of supervision, backbone models, and the scale and nature of training data.
We include three \textit{specialized} baselines, which are trained for the VPR task on large-scale urban datasets, and three new baselines that use the \texttt{CLS} token of the foundation models, as summarized in \cref{tab:baselines}.

\begin{table}[!t]
\centering
\caption{State-of-the-art Baselines used for Comparison}
\scalebox{0.8}{
\begin{tabular}{@{}llll@{}}
\toprule

\textbf{Method} & \textbf{Backbone} & \textbf{Training Dataset} & \textbf{Supervision} \\

\cmidrule{1-1} \cmidrule(lr{0.75em}){2-2} \cmidrule(lr{0.75em}){3-3} \cmidrule{4-4}

NetVLAD~\cite{arandjelovic2016netvlad, berton2022deep} & ResNet-18 & Pitts-30k & VPR - Contrastive \\

CosPlace~\cite{berton2022rethinking} & ResNet-101 & SF-XL & VPR - Classification \\

MixVPR~\cite{ali2023mixvpr} & ResNet-50 & GSV-Cities & VPR - Contrastive \\

\cdashmidrule{1-4}

CLIP~\cite{radford2021learning, ilharco2021openclip} & ViT-bigG-14 & Laion 2B & Image-Caption Pairs \\

\dino{}~\cite{caron2021emerging} & ViT-S8 & ImageNet & Self-Supervised \\

\dinovtwo{}~\cite{oquab2023dinov2} & ViT-G14 & LVD-142M & Self-Supervised \\

\bottomrule
\end{tabular}
}
\label{tab:baselines}
\end{table}

\subsubsection{\coolname{} - Nomenclature and Model Specifications}
\label{sec:specs}
All names are of the form \coolname{}-\texttt{aggregation}-\texttt{model}, where \texttt{aggregation} is one of \emph{VLAD}, \emph{GeM}; and \texttt{model} is one of \emph{DINO}, \emph{DINOv2}.
For \coolagg{VLAD}{\dino{}}, we use the ViT-S8 layer $9$ \texttt{key} facet features and 128 clusters for VLAD. Likewise, for \coolaggshort{GeM} and \coolagg{VLAD}{\dinovtwo{}}, we use ViT-G14 layer $31$ \texttt{value} facet features, with 32 clusters for VLAD.

\begin{table*}[!t]
\centering
\caption{Performance comparison on Benchmark Structured Environments}
\scalebox{0.95}{
\begin{tabular}{@{}lcccccccccccccc@{}}
\toprule

& \multicolumn{2}{c}{\color{IndoorDark} \indoorChar \hspace{0.1 em} \viewpointChar}  & \multicolumn{2}{c}{\color{IndoorDark} \indoorChar \hspace{0.1 em} \lightingChar} & \multicolumn{2}{c}{\color{IndoorDark} \indoorChar \hspace{0.1 em} \lightingChar}   & \multicolumn{2}{c}{\color{OutdoorDark} \outdoorChar \hspace{0.1 em} \viewpointChar}   & \multicolumn{2}{c}{\color{OutdoorDark} \outdoorChar}  & \multicolumn{2}{c}{\color{OutdoorDark} \outdoorChar \hspace{0.1 em} \lightingChar}    & \multicolumn{2}{c}{}     \\ 

& \multicolumn{2}{c}{\color{IndoorDark} \textbf{Baidu Mall}}  & \multicolumn{2}{c}{\color{IndoorDark} \textbf{Gardens Point}} & \multicolumn{2}{c}{\color{IndoorDark} \textbf{17 Places}}   & \multicolumn{2}{c}{\color{OutdoorDark} \textbf{Pitts-30k}}   & \multicolumn{2}{c}{\color{OutdoorDark} \textbf{St Lucia}}  & \multicolumn{2}{c}{\color{OutdoorDark} \textbf{Oxford}}    & \multicolumn{2}{c}{\textbf{Average}}     \\ 

\cmidrule(l){2-15} 

\textbf{Methods}                     & R@1            & R@5           & R@1              & R@5           & R@1            & R@5           & R@1            & R@5           & R@1            & R@5         & R@1            & R@5         & R@1            & R@5        \\ 

\cmidrule{1-1} \cmidrule(lr{0.75em}){2-3} \cmidrule(lr{0.75em}){4-5} \cmidrule(lr{0.75em}){6-7} \cmidrule(lr{0.75em}){8-9} \cmidrule(lr{0.75em}){10-11} \cmidrule(lr{0.75em}){12-13} \cmidrule{14-15}

NetVLAD~\cite{arandjelovic2016netvlad}  & 53.1          & 70.5          & 58.5            & 85.0          & 61.6          & 77.8          & 86.1          & 92.7          & 57.9          & 73.0        & 57.6          & 79.1        &    62.5       &  79.7         \\

CosPlace~\cite{berton2022rethinking} & 41.6          & 55.0          & 74.0            & 94.5          & 61.1          & 76.1          & 90.4          & \textbf{95.7}          & 99.6 & 99.9          & 95.3 & 99.5 & 77.0          & 86.8          \\

MixVPR~\cite{ali2023mixvpr}   & 64.4 & 80.3          & 91.5            & 96.0          & 63.8          & 78.8          & \textbf{91.5} & 95.5 & \textbf{99.7}          & \textbf{100} & 92.7          & 99.5          & 83.9 & 91.7 \\

\cdashmidrule{1-15}

CLIP-\texttt{CLS}~\cite{radford2021learning}  & 56.0          & 71.6          & 42.5            & 74.5          & 59.4          & 77.6 & 55.0          & 77.2         & 62.7          & 80.7        & 46.6          & 60.7        & 53.7    & 73.7          \\

\dino{}-\texttt{CLS}~\cite{caron2021emerging} & 48.3 & 65.1 & 78.5 & 95.0 & 61.8 & 76.4 & 70.1 & 86.4 & 45.2 & 64.0 & 20.4 & 46.6 & 54.1 & 72.3 \\

\dinovtwo{}-\texttt{CLS}~\cite{oquab2023dinov2} & 49.2 & 64.6 & 71.5 & 96.0 & 61.8 & 78.8 & 78.3 & 91.1 & 78.6 & 89.7 & 47.1 & 58.1 & 64.4 & 79.7 \\

\rowcolor{Light}
\coolagg{GeM}{\dinovtwo{}} & 50.1 & 70.6 & 88.0 & 97.5 & 63.6 & 79.6 & 77.0 & 87.3 & 76.9 & 89.3 & 92.2 & 97.9 & 74.6 & 87.0
\\

\rowcolor{Light}
\coolagg{VLAD}{\dino{}} & 61.2 & 78.3 & 95.0 & 98.5 & 63.8 & 78.8 & 83.4 & 92.0 & 88.5 & 94.9 & 82.2 & 99.0 & 79.0 & 90.2
\\

\rowcolor{Light}
\coolagg{VLAD}{\dino{}}\textit{-PCA} & 62.3 & 81.2 & 91.5 & 99.5 & 63.3 & 78.8 & 82.8 & 90.8 & 87.6 & 94.3 & 82.7 & 96.3 & 78.4 & 90.1
\\

\rowcolor{Light}
\textbf{\coolagg{VLAD}{\dinovtwo{}}}   & \textbf{75.2}  & 87.6 & 95.5 & \textbf{99.5} & \textbf{65.0} & 80.5 & 87.7 & 94.7 & 96.2 & 98.8 & \textbf{99.5} & \textbf{100} & \textbf{86.5} & 93.5
\\

\rowcolor{Light}
\textbf{\coolagg{VLAD}{\dinovtwo{}}\textit{-PCA}}   & 74.9  & \textbf{89.4} & \textbf{96.0} & \textbf{99.5} & 64.8 & \textbf{81.0} & 86.9 & 93.8 & 96.4 & 99.5 & 96.9 & \textbf{100} & 86.0 & \textbf{93.9}
\\

\bottomrule
\end{tabular}
}
\label{tab:mainResult}
\end{table*}

\begin{table*}[!t]
\centering
\caption{Performance Comparison on Unstructured Environments}
\scalebox{0.95}{
\begin{tabular}{@{}lcccccccccccccc@{}}
\toprule

& \multicolumn{2}{c}{\color{SubTDark} \hawkinsChar \hspace{0.1 em} \oppositeChar} & \multicolumn{2}{c}{\color{SubTDark} \subtChar \hspace{0.1 em} \oppositeChar} & \multicolumn{2}{c}{\color{AerialDark} \aerialChar \hspace{0.1 em} \oppositeChar} & \multicolumn{2}{c}{\color{AerialDark} \aerialChar} & \multicolumn{2}{c}{\color{AerialDark} \aerialChar \hspace{0.1 em} \viewpointChar} & \multicolumn{2}{c}{\color{UnderWaterDark} \underwaterChar} & \multicolumn{2}{c}{}\\

& \multicolumn{2}{c}{\color{SubTDark} \textbf{Hawkins}} & \multicolumn{2}{c}{\color{SubTDark} \textbf{Laurel Caverns}} & \multicolumn{2}{c}{\color{AerialDark} \textbf{Nardo-Air}} & \multicolumn{2}{c}{\color{AerialDark} \textbf{Nardo-Air R}} & \multicolumn{2}{c}{\color{AerialDark} \textbf{VP-Air}} & \multicolumn{2}{c}{\color{UnderWaterDark} \textbf{Mid-Atlantic Ridge}} & \multicolumn{2}{c}{\textbf{Average}}\\ 

\cmidrule(l){2-15}

\textbf{Methods} & R@1 & R@5 & R@1 & R@5 & R@1 & R@5 & R@1 & R@5 & R@1 & R@5 & R@1 & R@5 & R@1 & R@5 \\ 

\cmidrule{1-1} \cmidrule(lr{0.75em}){2-3} \cmidrule(lr{0.75em}){4-5} \cmidrule(lr{0.75em}){6-7} \cmidrule(lr{0.75em}){8-9} \cmidrule(lr{0.75em}){10-11} \cmidrule(lr{0.75em}){12-13} \cmidrule{14-15}

NetVLAD~\cite{arandjelovic2016netvlad} & 34.8 & 71.2 & 39.3 & 71.4 & 19.7 & 39.4 & 60.6 & 85.9 & 6.4 & 17.7 & 25.7 & 53.5 & 31.1 & 56.5 \\

CosPlace~\cite{berton2022rethinking}   & 31.4 & 59.3 & 24.1 & 47.3 & 0 & 1.4 & 91.6 & \textbf{100} & 8.1 & 14.2 & 20.8 & 40.6 & 29.3 & 43.8 \\

MixVPR~\cite{ali2023mixvpr}   & 25.4 & 60.2 & 29.5 & 67.0 & 32.4 & 42.2 & 76.1 & 98.6 & 10.3 & 18.3 & 25.7 & 60.4 & 33.2 & 57.8 \\

\cdashmidrule{1-15}
CLIP-\texttt{CLS}~\cite{radford2021learning}  & 33.0 & 67.0 & 36.6 & 66.1 & 42.2 & 70.4 & 62.0 & 97.2 & 36.6 & 52.8 & 25.7 & 51.5 & 39.4 & 67.5 \\

\dino{}-\texttt{CLS}~\cite{caron2021emerging} & 46.6 & 84.8 & 41.1 & 57.1 & 57.8 & 90.1 & 84.5 & \textbf{100} & 24.0 & 38.4 & 27.7 & 49.5 & 47.0 & 70.0 \\

\dinovtwo{}-\texttt{CLS}~\cite{oquab2023dinov2} & 28.0 & 62.7 & 40.2 & 65.2 & 73.2 & 88.7 & 71.8 & 91.6 & 45.2 & 59.9 & 24.8 & 48.5 & 47.2 & 69.4 \\

\rowcolor{Light}
\coolagg{GeM}{\dinovtwo{}} & 53.4 & 83.9 & 58.9 & 86.6 & \textbf{76.1} & 83.1 & 57.8 & 97.2 & 38.3 & 53.8 & 14.8 & 49.5 & 49.9 & 75.7
\\

\rowcolor{Light}
\coolagg{VLAD}{\dino{}}  & 48.3 & 84.8 & 57.1 & 79.5 & 43.7 & 54.9 & \textbf{94.4} & \textbf{100} & 17.8 & 28.7 & \textbf{41.6} & \textbf{66.3} & 50.5 & 69.0 
\\

\rowcolor{Light}
\textbf{\coolagg{VLAD}{\dinovtwo{}}}   & \textbf{65.2} & \textbf{94.1} & \textbf{61.6} & \textbf{90.2} & \textbf{76.1} & \textbf{94.4} & 85.9 & \textbf{100} & \textbf{66.7} & \textbf{79.2} & 34.6 & 61.4 & \textbf{65.0} & \textbf{86.5}
\\

\bottomrule
\end{tabular}
}
\label{tab:unstructured}
\end{table*}

\section{Experiments, Results, and Analyses}
\label{sec:results}

We first evaluate \coolname{} against SOTA VPR techniques and report results across structured \& unstructured environments, viewpoint shifts, and temporal appearance variations.
We further present a comparative analysis of the specialized baselines and variants directly using the \texttt{CLS} token (i.e., per-image features).
We then present a detailed vocabulary analysis followed by insights into the design of \coolname{}.
Lastly, we demonstrate the benefits of self-supervised ViTs by contrasting them with existing VPR-trained ViTs.

\subsection{State-of-the-art Comparison}

\subsubsection{Structured Environments}

\cref{tab:mainResult} highlights the \highlight{general applicability of the \coolname{} methods on structured environments}, in particular, the \indoor{} and \urban{} domains.
\coolagg{VLAD}{\dinovtwo{}} achieves the highest recall across all the \indoor{} datasets while outperforming MixVPR (the second best) and CosPlace by 5\% and 20\% on average (R@1).
Interestingly, foundation models' \texttt{CLS} descriptors (while being inferior to our method) are competitive with baselines such as CosPlace and NetVLAD, e.g., CLIP outperforms them respectively by 15\% and 3\% on Baidu Mall.
Through our proposed use of feature aggregation for foundation models, we observe that simply using GeM pooling over \dinovtwo{} features (i.e., \coolagg{GeM}{\dinovtwo{}}) significantly improves performance over the \dinovtwo{} \texttt{CLS} token.
This is further improved by \coolaggshort{VLAD}, which beats all prior approaches on these datasets.
In the \urban{} case -- which well aligns with the training distribution of the baselines supervised specifically for VPR on urban data -- we observe that \coolaggshort{VLAD} is inferior by 3-4\% on daytime conditions of Pitts30k and St Lucia, but it achieves state-of-the-art for day-night variations on Oxford.
We further showcase that a PCA-Whitening of the \coolaggshort{VLAD} descriptors using the domain-specific database enables similar SOTA performance while having a $100 \times$ smaller embedding size ($49k$ to $512$).

\subsubsection{Unstructured Environments}

\highlight{\cref{tab:unstructured} highlights the fragility of the specialized baselines and shows that \coolname{} outperforms all the baselines by a \textit{large} margin in these challenging unstructured environments}. 
Even the \texttt{CLS} methods outperform VPR-specialized baselines, e.g., \dinovtwo{}-\texttt{CLS} exceeds MixVPR by 41\% on Nardo-Air and 35\% on VP-Air under strong viewpoint variations.
The \coolname{} methods consistently outperform both the specialized and the \texttt{CLS} baselines, where the best performers in the respective categories, i.e.,  MixVPR and \dinovtwo{}-\texttt{CLS}, lag behind \coolaggshort{VLAD} by 32\% and 18\% on average (R@1).

\subsubsection{Temporal \& Viewpoint Changes} 

We further demonstrate the \highlight{robustness of \coolname{} for \textit{anytime} and \textit{anyview} VPR}.
We evaluate multiple datasets where revisiting a place at different time intervals leads to variations in scene appearance (\textit{anytime}).
In comparison to the SOTA VPR baselines, MixVPR/CosPlace, we observe the following gains using \coolaggshort{VLAD} on different temporal changes: {5/11}\% on day-night cycles affecting outdoors (Oxford), indoors (17 Places), and mixture (Gardens Point); {9/8}\% on seasonal shifts (Oxford); {21/28}\% on long period jumps (2022 vs. 2023 for Nardo-Air, 2015 Vs. 2020 for the Mid-Atlantic Ridge).
A similar trend is observed for viewpoint shifts (\textit{anyview}), where we test on datasets that vary both in terms of the \textit{view-type}, e.g., street vs aerial, and the \textit{shift-type}. 
\coolaggshort{VLAD} outperforms MixVPR/CosPlace on orientation-based shifts by {21/30}\% and extreme 90$^\circ$/180$^\circ$ shifts by {39/49}\%.

\subsubsection{Specialized Baselines} 

The average recall of NetVLAD, CosPlace, and MixVPR confirms the \highlight{general trend of better performance in task-specific baselines with an increasing scale of urban training data}, combined with innovations in learning objective (CosPlace) and learnable aggregation (MixVPR).
Additionally, we observe one peculiar failure case of CosPlace on the Nardo-Air dataset.
No correct matches were found under the combined effect of out-of-distribution (aerial) and extreme viewpoint (90 degrees) shifts.
Visual inspection revealed that all queries incorrectly matched to a handful of reference images having similar orientation of fields and roads.

\subsubsection{\texttt{CLS} vs. Aggregation (\coolname{})} 

\highlight{When the foundation models are used with local feature aggregation instead of the \texttt{CLS} token, we observe significant performance jumps}: \dinovtwo{}-based \coolaggshort{GeM} and \coolaggshort{VLAD} outperform \dinovtwo{}-\texttt{CLS} by 9\%/2\% and 23\%/18\% respectively on structured/unstructured environments.
Furthermore, the average recall of the \texttt{CLS} token-based global descriptors (CLIP, \dino{} \& \dinovtwo{}) indicates their superiority to specialized baselines on unstructured environments.

\begin{table}[!t]
\centering
\caption{Effect of vocabulary type on R@1 for \coolagg{VLAD}{\dinovtwo}}
\scalebox{1}{
\begin{tabular}{@{}lccc@{}}
\toprule

& \color{IndoorDark} \indoorChar  & \color{OutdoorDark} \outdoorChar & \color{AerialDark} \aerialChar \\ 

\textbf{Vocabulary Type} & \color{IndoorDark} \textbf{Indoor}  & \color{OutdoorDark} \textbf{Urban} & \color{AerialDark} \textbf{Aerial} \\ 

\cmidrule{1-1} \cmidrule(lr{0.75em}){2-2} \cmidrule(lr{0.75em}){3-3} \cmidrule{4-4}

Global & 77.0 & 93.9 & 57.1 \\

Structured & 77.0 & 93.3 & 56.4 \\

Unstructured & 74.8 & 89.0 & 75.8 \\

Map-Specific & 78.0 & 92.3 & 62.9 \\

Domain-Specific  & \textbf{78.6} & \textbf{94.4} & \textbf{76.2} \\

\bottomrule
\end{tabular}
}
\label{tab:vocab}
\end{table}

\subsection{Vocabulary Analysis}

\subsubsection{Vocabulary Source}
\label{sec:vocab_ablations}

\highlight{\cref{tab:vocab} shows how the vocabulary source used for VLAD influences recall, where domain-specific vocabulary leads to the best recall.}
We construct multiple VLAD vocabularies using different subsets of the 12 datasets used in this work and report average recall per domain.
As described in \cref{sec:vocab_approach}, the subsets for different domains are obtained through a qualitative PCA visualization (see \cref{fig:splash}), which is quantitatively justified through the results presented here.
The other vocabulary sources that we compare against are: \textit{Global} using all 12 datasets; \textit{Structured} using 3 indoor and 3 urban datasets; \textit{Unstructured} using the complement set of structured; and \textit{Map-specific} using only the reference database of a particular dataset.
In the aerial domain, domain-specific achieves 13\% over map-specific and 19\% over global vocabulary.

\begin{table}[!t]
\centering
\caption{Analysing intra-domain transferability of \coolagg{VLAD}{\dinovtwo{}} vocabularies}
\scalebox{0.85}{
\begin{tabular}{@{}llcc@{}}
\toprule

\textbf{Vocabulary} & \textbf{Evaluation} & \textbf{Map-Specific} &  \textbf{Vocab-Transfer} \\

\textbf{Dataset} & \textbf{Dataset} & \textbf{Recall@1} & \textbf{Recall@1} \\

\cmidrule{1-1} \cmidrule(lr{0.75em}){2-2} \cmidrule(lr{0.75em}){3-3} \cmidrule{4-4}

\color{IndoorDark} \textbf{Baidu Mall (0.7k)} & \color{IndoorDark} \textbf{17 Places (0.4k)} & \textbf{64.5} & 63.8  \\
& \color{IndoorDark} \textbf{Gardens Point (0.2k)} & \textbf{98.0} & 94.5  \\

\midrule

\color{AerialDark} \textbf{VP-Air (2.7k)} & \color{AerialDark} \textbf{Nardo-Air (0.1k)} & 57.8 & \textbf{64.8}  \\
& \color{AerialDark} \textbf{Nardo-Air R (0.1k)} & 70.4 & \textbf{88.7}  \\

\midrule

\color{OutdoorDark} \textbf{Pitts-30k (10k)} & \color{OutdoorDark} \textbf{Oxford (0.2k)} & 94.8 & \textbf{99.0}\\

\bottomrule
\end{tabular}
}
\label{tab:transfer}
\end{table}

\begin{figure}[!t]
\centering
\begin{tabular}{cccc}
\includegraphics[trim={0cm 1.2cm 7.5cm 0cm},clip,width=0.88\linewidth]{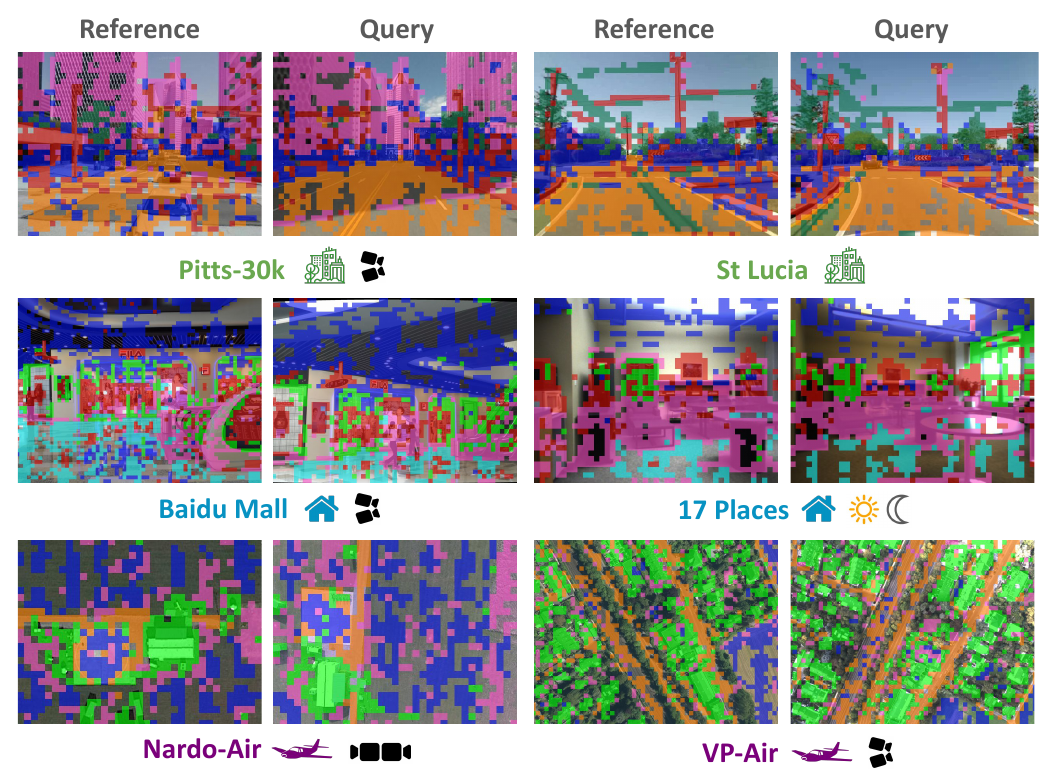} 
\end{tabular}
\caption{VLAD cluster assignment visualizations of the reference-query pairs highlight the \textbf{intra-domain consistency} of the domain-specific vocabulary. Similar colors across images of a specific domain indicate matched clusters.}
\label{fig:cluster_viz}
\end{figure}

\subsubsection{Consistency}

\cref{fig:cluster_viz} showcases the \highlight{robust intra-domain consistency of the domain-specific vocabulary}, further justifying the high performance of \coolaggshort{VLAD}.
Specifically, we visualize the cluster assignments (with $K = 8$) for the local features using the domain-specific vocabulary.
In the \urban{} domain, the roads, pavements, buildings, and vegetation are consistently assigned to the same cluster across changing conditions and places. 
For the \indoor{} domain, we can observe intra-domain consistency for the floor \& ceiling, while there is intra-place consistency for the text signs and furniture.
For the \aerial{} domain, it can be observed that roads, vegetation, and buildings are assigned to unique clusters across both the rural and urban images.

\begin{figure*}[!t]
\centering
\begin{tabular}{ccccc}
\includegraphics[trim={0.2cm 0cm 0.2cm 0cm}, clip, width=0.165\linewidth]{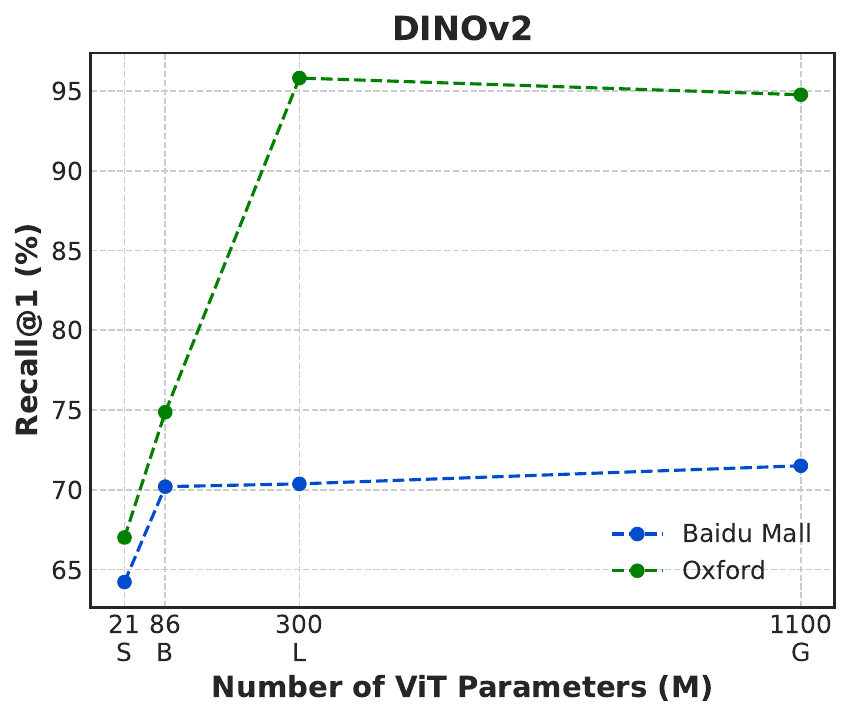} 
&
\includegraphics[trim={0.2cm 0cm 0.2cm 0cm}, clip, width=0.176\linewidth]{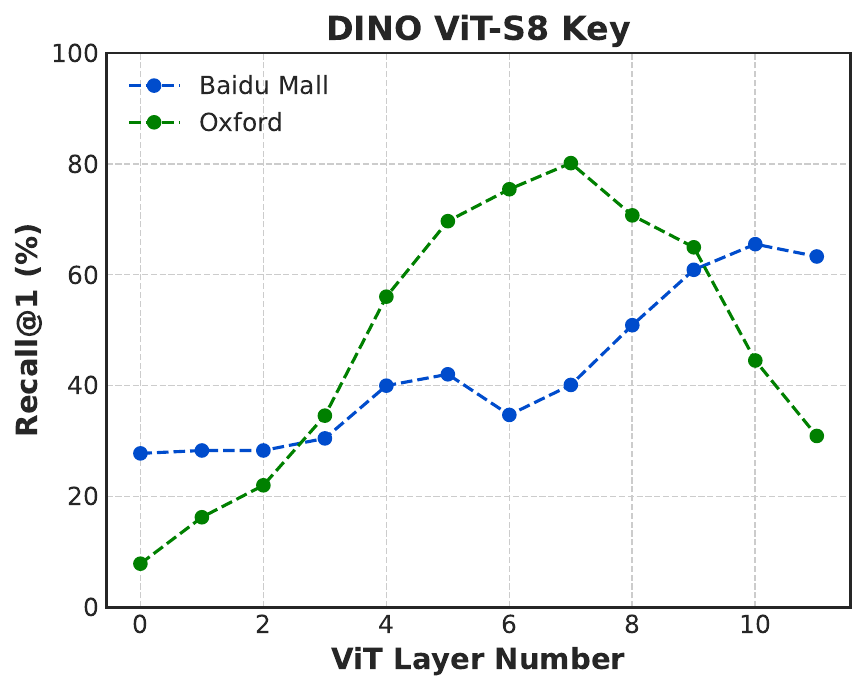} 
&
\includegraphics[trim={0.2cm 0cm 0.2cm 0cm}, clip, width=0.176\linewidth]{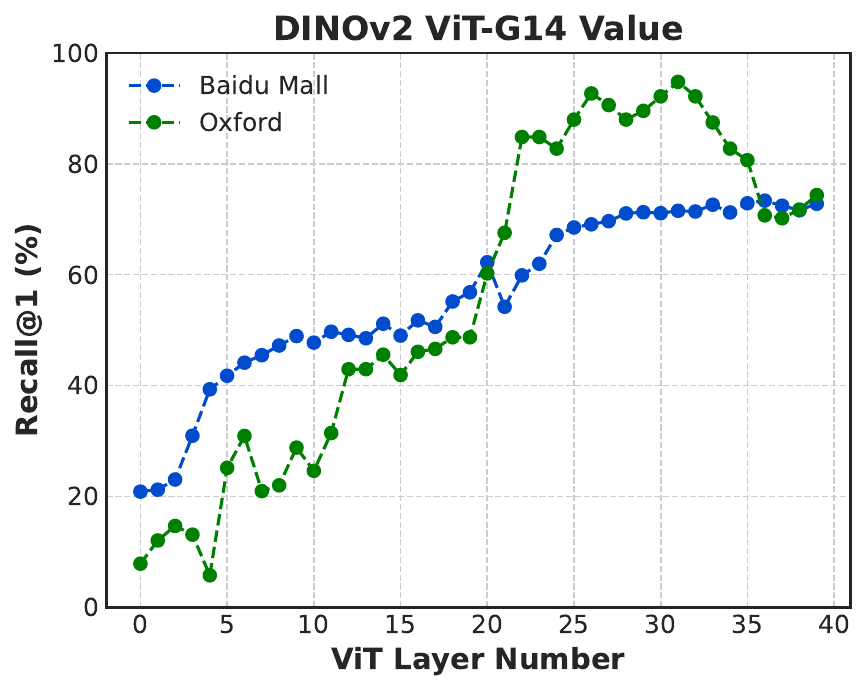} 
&
\includegraphics[trim={1cm -0.2cm 0 0}, width=0.177\linewidth]{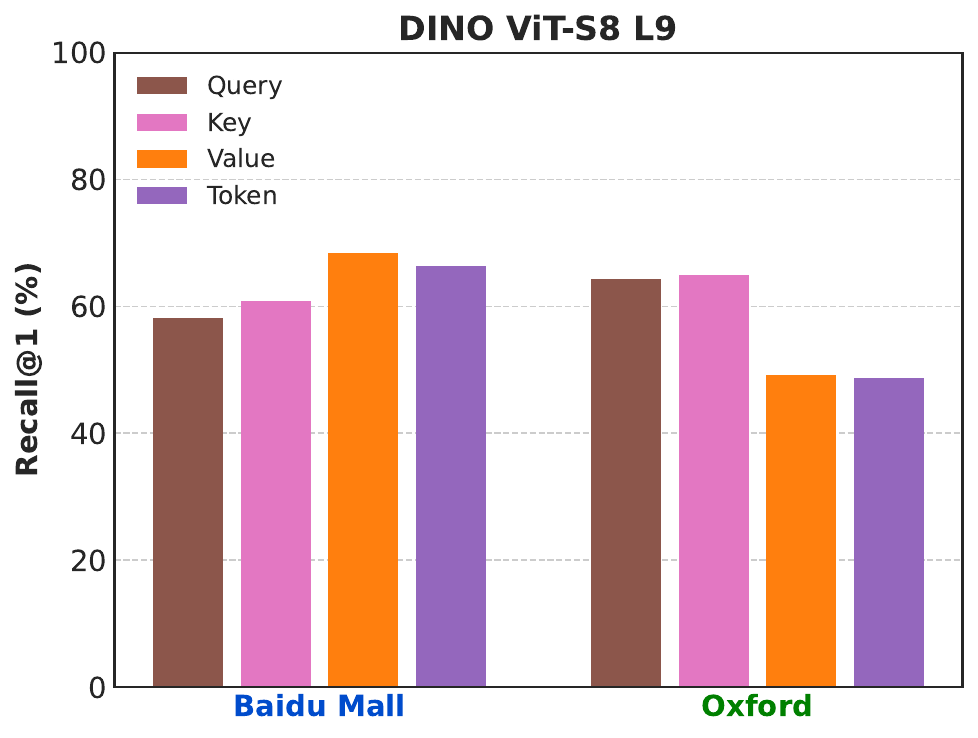}
&
\includegraphics[trim={1cm -0.2cm 0 0}, width=0.177\linewidth]{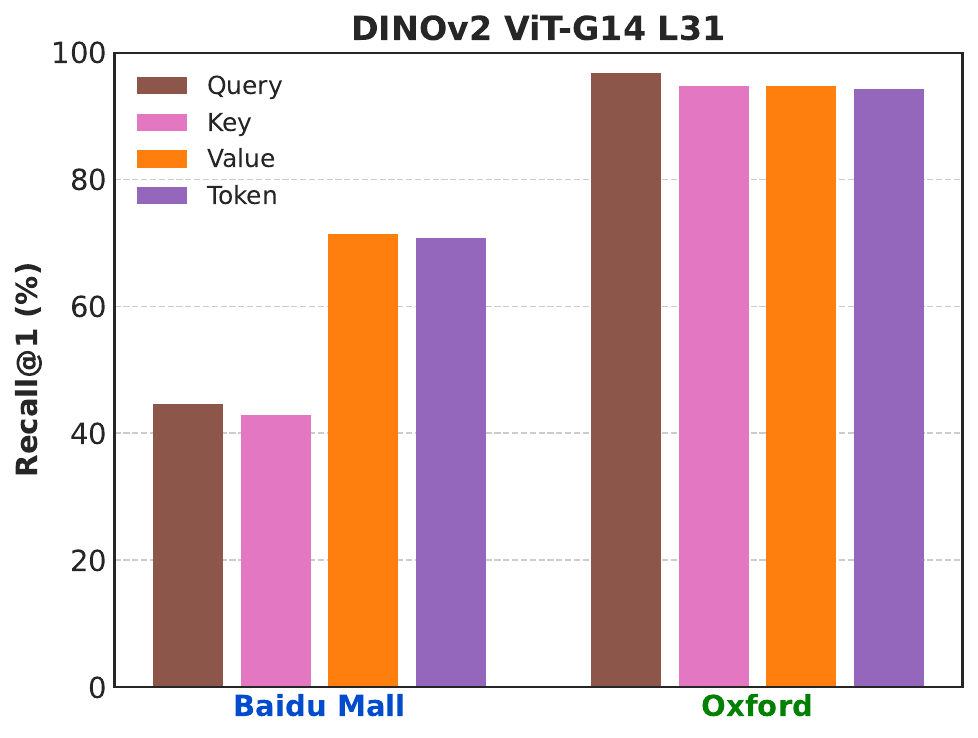} 
\\
(a) Model & \multicolumn{2}{c}{(b) Layer} & \multicolumn{2}{c}{(c) Facet}
\end{tabular}
\caption{\textbf{Design Choices for \coolaggshort{VLAD}}: (a) Performance scales with the model size but saturates at ViT-L.
(b) Performance peaks at intermediate layers instead of the final layer for both \dino{} \& \dinovtwo{}. (c) On average, \texttt{key} \& \texttt{value} perform the best respectively for \dino{} \& \dinovtwo{}.}
\label{fig:ablations}
\end{figure*}

We further demonstrate that this \highlight{robust consistency within a domain enables us to deploy \coolaggshort{VLAD} in target environments with small reference databases (maps) that lack information richness}.
For datasets belonging to a given domain, we pick the largest reference database to form the vocabulary and evaluate on other datasets from that domain. 
In \cref{tab:transfer}, for \aerial{} and \urban{} domains, we can observe that 7-18\% higher R@1 can be achieved when using a larger source of vocabulary as compared to just using the target dataset's own smaller map, thus demonstrating the transferability of vocabularies within the same domain. 
For the \indoor{} domain, the drop in performance is either due to a relatively limited size of the largest reference database or the large diversity across datasets, e.g., shops in Baidu Mall compared to offices in the other two datasets. Nevertheless, when using this unified diverse vocabulary from all the datasets in the indoor domain, the overall recall is better than using map-specific vocabularies, as shown in~\cref{tab:vocab}.

\subsection{Insights into \coolname{} Design}
We present insights on varying parameters within \coolname{}, using two datasets, Baidu Mall \& Oxford, which are representative of the typical VPR challenges:

\subsubsection{ViT Architecture}

\subfigref[a]{ablations} showcases that \highlight{larger \dinovtwo{} ViT backbones lead to better performance}, where the performance tends to saturate at ViT-L (300 million parameters). 
Since, on average, ViT-G performs better than ViT-L, we use ViT-G for \dinovtwo{}.
For \dino{}, we use ViT-S, which is the only available architecture.

\subsubsection{ViT Layers \& Facets}
\label{sec:facets_and_layers}

\subfigref[b]{ablations} shows that \highlight{peak performance is achieved through deeper layers, somewhere between the middle and the last layer}. 
For a smaller ViT architecture (\dino{} ViT-S on the left), it can be observed that middle layers have higher performance on Oxford.
This can be attributed to their higher positional encoding bias, which is helpful under no viewpoint shift across reference-query pairs. 
Hence, aligning with the findings presented in \cref{sec:features}, we choose $9$ and $31$ as our operating layers for \dino{} and \dinovtwo{}, respectively.

In \subfigref[c]{ablations}, the \highlight{\texttt{key} \& \texttt{value} facets consistently achieve high recall for \dino{} \& \dinovtwo{} respectively}.
Although \texttt{query} and \texttt{key} facets perform better on Oxford when using \dino{} (left), this gap diminishes when using \dinovtwo{} (right).  
The performance difference between the \texttt{query} \& \texttt{value} gets inverted from Baidu to Oxford; indicating a high positional bias in the \texttt{query} \& \texttt{key}, leading to poor performance under the significant viewpoint shift in Baidu.

\begin{table}[!t]
\centering
\caption{Analysis comparing the Recall@1 \& Descriptor Dimensionality across varying aggregation methods}
\scalebox{0.7}{
\begin{tabular}{@{}lcccccc@{}}
\toprule

& \multicolumn{3}{c}{\textbf{\dino{}}} & \multicolumn{3}{c}{\textbf{\dinovtwo{}}} \\

\cmidrule(lr{0.75em}){2-4} \cmidrule{5-7}

\textbf{Aggregation Methods}  & \color{IndoorDark} \textbf{Baidu} $\uparrow$ & \color{OutdoorDark} \textbf{Oxford} $\uparrow$ & \textbf{Dim} $\downarrow$ & \color{IndoorDark} \textbf{Baidu} $\uparrow$ & \color{OutdoorDark} \textbf{Oxford} $\uparrow$ & \textbf{Dim} $\downarrow$ \\

\cmidrule{1-1} \cmidrule(lr{0.75em}){2-4} \cmidrule{5-7}

Global Average Pool (GAP) & 29.6 & 28.8 & \textbf{384} & 41.6 & 78.5 & \textbf{1536} \\

Global Max Pool (GMP)  & 34.9 & 38.2 & \textbf{384} & 64.4 & 74.9 & \textbf{1536} \\

Generalized Mean Pool (GeM)  & 34.7 & 47.6 & \textbf{384} & 50.1 & 92.2 & \textbf{1536} \\

Soft Assignment VLAD  & 33.8 & 28.3 & 49152 & 40.3 & 82.2 & 49152 \\

Hard Assignment VLAD & \textbf{60.9} & \textbf{64.9} & 49152 & \textbf{71.5} & \textbf{94.8} & 49152 \\

\bottomrule
\end{tabular}
}
\label{tab:aggregation}
\end{table}

\subsubsection{Aggregation Methods}

In \cref{tab:aggregation}, we compare the various unsupervised local feature aggregation techniques as discussed in \cref{sec:aggregating_features} and observe that \highlight{hard assignment-based VLAD works the best}.
We can further see that the vocabulary-free methods provide an optimal trade-off between performance and storage, where GeM pooling tends to do the best.
Also, we observed that hard assignment is typically $1.4$ times faster than soft assignment.

\begin{table}[!t]
\centering
\caption{Analysis comparing the Recall@1 of VPR-trained ViTs to Self-supervised ViTs}
\scalebox{0.78}{
\begin{tabular}{@{}lccccc@{}}
\toprule

& \color{IndoorDark} \indoorChar  & \color{OutdoorDark} \outdoorChar & \color{AerialDark} \aerialChar & \color{SubTDark} \subtChar \hspace{0.1 em} \hawkinsChar  & \color{UnderWaterDark} \underwaterChar\\ 

\textbf{Method} & \color{IndoorDark} \textbf{Indoor}  & \color{OutdoorDark} \textbf{Urban} & \color{AerialDark} \textbf{Aerial} & \color{SubTDark} \textbf{SubT \& D} & \color{UnderWaterDark} \textbf{Underwater} \\ 

\cmidrule{1-1} \cmidrule(lr{0.75em}){2-2} \cmidrule(lr{0.75em}){3-3} \cmidrule(lr{0.75em}){4-4} \cmidrule(lr{0.75em}){5-5} \cmidrule{6-6}

ViT-B CosPlace & 62.9 & 80.7 & 26.3 & 26.5 & 18.8 \\

ViT-B CosPlace-VLAD & 68.5 & 82.9 & 38.4 & 37.5 & 23.8 \\

\rowcolor{Light}
ViT-S \coolagg{VLAD}{\dino{}} & 72.9 & 79.6 & 47.8 & 52.7 & \textbf{41.6} \\

\rowcolor{Light}
ViT-B \coolagg{VLAD}{\dinovtwo{}} & 77.0 & 82.6 & 53.6 & 60.2 & 35.6 \\

\rowcolor{Light}
ViT-G {\coolagg{VLAD}{\dinovtwo{}}} & \textbf{78.0} & \textbf{92.3} & \textbf{62.9} & \textbf{63.4} & 34.6 \\

\bottomrule
\end{tabular}
}
\label{tab:vit}
\end{table}

\subsection{Self-supervised vs VPR-supervised ViT}

\cref{tab:vit} shows that the \highlight{high performance of \coolaggshort{VLAD} is not a consequence of simply using a large ViT but an outcome of self-supervised training on large-scale curated data, which leads to generality in the underlying features}~\cite{oquab2023dinov2}.
In particular, we compare a ViT trained specifically for VPR (i.e., CosPlace~\cite{berton2022rethinking}) against those based on self-supervision (i.e., \dino{} \& \dinovtwo{}).
For the VPR-supervised CosPlace, we include the authors' GeM pooling-based ViT-B model along with its adapted version that uses a VLAD layer ($K = 128$) on top of ViT-B's $6$th layer (which performed better than other layers). 
For self-supervised methods, we include \coolaggshort{VLAD} variants: \dino{}~ViT-S, \dinovtwo{}~ViT-B and ViT-G. 
All VLAD-based methods in these comparisons use map-specific vocabulary.
Comparing ViT-B-based methods, we can observe that even though CosPlace's overall performance improves with VLAD, \coolagg{VLAD}{\dinovtwo{}} outperforms it by 8-13\%.
Interestingly, even ViT-S based \coolagg{VLAD}{\dino{}} outperforms ViT-B-based CosPlace-VLAD by 4-18\% while using 4$\times$ fewer parameters.
The only exception to these trends is in the urban domain, where CosPlace-VLAD outperforms ViT-S and ViT-B based \coolaggshort{VLAD}, which is justified by CosPlace's VPR-specific training on urban data.
Despite this, \coolagg{VLAD}{\dinovtwo{}} ViT-G surpasses all other methods.

\section{Conclusion}

This paper introduces \coolname{} -- a significant step towards \emph{universal} VPR. Driven by the limitations of \textit{environment-} and \textit{task-specific} VPR techniques, and the fragility of per-image features extracted from foundation models, we propose to blend the per-pixel features computed by these models with unsupervised feature aggregation techniques like VLAD and GeM.
Through our benchmarking and analyses on a diverse suite of datasets, we shed light on the brittleness of current large-scale urban-trained VPR approaches and show that \coolname{} outperforms the previous state-of-the-art by up to $4\times$.
This work stretches the applicability scope of VPR and, in turn, robot localization to \textit{anytime}, \textit{anywhere} \& under \textit{anyview}, which is crucial to enable downstream capabilities, such as robot navigation in the wild.

\section*{Acknowledgments}

This work was supported by ARL grant W911QX20D0008/W911QX22F0078(TO6).
The authors thank Ivan Cisneros \& Yao He for collecting the Nardo-Air dataset.
We also thank the members of CMU AirLab for their insightful discussions throughout this project.
\ifarxiv
Lastly, the authors thank Deepak Pathak, Murtaza Dalal, Ananye Agarwal, Aditi Raghunathan, and Tuomas Sandholm for feedback on an initial version of the work.
Parts of this work used Bridges-2 at PSC through allocation cis220039p from the ACCESS program, which is supported by NSF grants 2138259, 2138286, 2138307, 2137603, and 213296.
\fi

\ifarxiv

\section*{Appendix}

\setcounter{section}{0}
\setcounter{equation}{0}
\setcounter{figure}{0}
\setcounter{table}{0}

\renewcommand{\thesection}{A\arabic{section}}
\renewcommand{\thesubsection}{A\arabic{subsection}}
\renewcommand{\thefigure}{A.\arabic{figure}}
\renewcommand{\thetable}{A.\arabic{table}}
\renewcommand{\theequation}{A.\arabic{equation}}

\section{Contribution statement}

\textbf{Nikhil Keetha} conceived the idea and led the project. Responsible for initial code development, writing major sections of the paper, and producing figures, tables \& videos.

\textbf{Avneesh Mishra} implemented vital components, including the foundation model feature extraction and modular scripts, to run experiments at a large scale. Responsible for running the ablation experiments, writing the first draft of the results section, and producing qualitative visualizations \& the Hugging Face demo.

\textbf{Jay Karhade} scaled the evaluation to a diverse suite of unstructured environments, implemented the vocabulary ablations, and performed explorations into various foundation models, including SAM. Responsible for the website, retrieval visualizations, and diverse suite of interactive demos.

\textbf{Krishna Murthy} was actively involved in brainstorming and critical review throughout the project. Responsible for the exploration of self-supervised visual foundation models. Wrote \& proofread sections of the paper.

\textbf{Sebastian Scherer} pushed us towards evaluating the practicality of current VPR systems in unstructured environments and developing a universal VPR system. Suggested a vital paper restructuring to ensure the critical message and insights are easily parsable. Sebastian provided compute resources for initial explorations and ablations.

\textbf{Madhava Krishna} was involved in initial brainstorming discussions and provided feedback throughout the development. Suggested revisions for sections of the paper. Madhav also provided most of the compute for the experiments conducted in this work.

\textbf{Sourav Garg} provided resourceful visual place recognition perspectives and critical thoughts in the brainstorming sessions, which led to clear insights into the applicability of foundation model features for VPR. Wrote and proofread sections of the paper.
\section{Vocabulary Specifications}
\label{sec:vocab_specs}

To ensure near-similar reference image frequencies across datasets, for the \urban{} vocabulary, we use all images from
Oxford \& St Lucia, but only every $4$th image for the larger Pitts-30k dataset.
For \aerial{}, we use the whole Nardo-Air database, but only every $2$nd image for VP-Air.
To generate the \indoor{}, \subt{}, \degraded{}, and \underwater{} vocabularies, we use all respective reference databases.
\section{Dataset Details}

In this section, we provide detailed descriptions of the 12 diverse datasets used for evaluation.

\subsection{Structured Environments}
\label{sec:appendix_structured}

\paragraph{Baidu Mall} This visual localization dataset consists of images captured within a mall with varying camera poses. The dataset provides groundtruth location and 3D pose of an image, making it suited for both 6-Degrees of Freedom (DoF) Localization and VPR testing. We use the entire dataset consisting of $2292$ query images \& $689$ reference images for evaluation. This mall dataset presents interesting and challenging properties, including perceptually aliased structures, distractors for VPR (such as people), and semantically rich information, such as billboards and signs.

\paragraph{Gardens Point} This dataset contains two traverses through the Gardens Point campus of Queensland University of Technology (QUT) captured at different times of the day, i.e., day and night. Both the database and query traverses contain $200$ images, respectively. The drastic lighting changes and transitions from indoor to outdoor scenarios make it a difficult VPR dataset.

\paragraph{17 Places} This indoor dataset consists of traverse collected within buildings at York University (Canada) and Coast Capri Hotel (British Columbia). The reference and query traverses consist of $406$ images. The high clutter, change in lighting conditions, and semantically rich information make this dataset interesting.

\paragraph{Pittsburgh-30k} This benchmark VPR dataset consists of images collected at various locations and poses throughout downtown Pittsburgh. We use the test split consisting of $10,000$ database images and $6816$ query images. This dataset is challenging due to the presence of drastic viewpoint shifts, a large variety of geometric structures such as buildings, and distractors such as cars and pedestrians.

\paragraph{St Lucia} This dataset consists of daytime traverses collected using a stereo camera pair on a car, where the traverses span a total distance of $9.5$ km. The reference traverse consists of $1549$ images, while the query traverse consists of $1464$ images. A large number of loop closure events, reverse traverses, shadows, and vegetation make this dataset challenging.

\paragraph{Oxford RobotCar} This dataset consists of Oxford City traverses, which showcase shifts in seasonal cycles and daylight. We use a subsampled version of the Overcast Summer and Autumn Night traverses, similar to HEAPUtil~\cite{keetha2021hierarchical}. The original traverses are subsampled with an approximate spacing of $5$ meters to obtain a total of $213$ frames in the summer traverse and $251$ frames in the autumn night traverse with a total distance spanning $1.5$ Km. This dataset presents a challenging shift in visual appearance caused by the time of day and seasonal shifts.

\subsection{Unstructured Environments}
\label{sec:appendix_unstructured}

\paragraph{Hawkins} This dataset is an indoor mapping of an abandoned multi-floor hospital in Pittsburgh, where it is particularly challenging due to long corridors with visually-degraded features~\cite{zhao2023subtmrs}. 
In particular, we use a long corridor spanning $282$ m with a localization radius of 8 m, where the database and query images are collected from 2 opposing viewpoints (forward \& backward direction). 
The database and query set contain $65$ and $101$ images, respectively.

\paragraph{Laurel Caverns} This subterranean dataset consists of images collected using a handheld payload~\cite{zhao2023subtmrs}. 
The low illumination scenarios and lack of rich visual features make this dataset particularly challenging. 
The opposing viewpoint of the database and query images adds additional complexity to the strong distribution shift.
We use a $102$ m trajectory with a localization radius of 8 m, where the database and query sets contain $141$ and $112$ images, respectively.

\paragraph{Nardo-Air} This is a GNSS-denied localization dataset collected using a $100^{\circ}$ FoV downward-facing camera on board a hexacopter flying at $10$ m/s and an altitude of $50$ m across a grass-strip runway named Nardo. The reference database comprises $102$ images obtained from a Google Maps TIF satellite image, while the query set contains $71$ drone-collected imagery. The perceptual aliasing at the end of the runway and non-typical vegetative features combined with a long time shift make this dataset challenging. The -R variant of this dataset indicates rotation where the drone imagery is rotated to match the satellite image orientation. We use a $700$ m trajectory spanning across a square kilometer area, where the localization radius is $60$ m.

\paragraph{VP-Air} This aerial VPR dataset consists of $2,706$ database-query image pairs and $10,000$ distractors collected at $300$ m altitude with a downward-facing camera on an aircraft~\cite{schleiss2022vpair}. The dataset spans over $100$ km, encompassing various challenging landscapes such as urban regions, farmlands, and forests. We use a localization radius of $3$ frames.

\paragraph{Mid-Atlantic Ridge} We construct this dataset using the overlapping sequences of an underwater visual localization dataset~\cite{boittiaux2022eiffel}. It presents OOD challenges including seabed objects, low illumination, and appearance shifts over a long time period (2015 vs. 2020). The dataset contains $65$ database images and $101$ query images, where the trajectory spans $18$ m and the localization radius is $0.3$ m.

\fi

\bibliographystyle{IEEEtran}
\bibliography{IEEEtranBST/IEEEabrv, root}

\end{document}